\documentclass[12pt]{article}
\usepackage{amsmath}
\usepackage{graphicx}
\usepackage{cite}
\usepackage{tikz}
\usepackage{authblk}  % 用于处理作者和机构信息
\usepackage[a4paper,margin=2.5cm]{geometry}
\usepackage{hyperref}  % 只加载一次

\title{Watertox: The Art of Simplicity in Universal Attacks \\
A Cross-Model Framework for Robust Adversarial Generation}

\author[1]{\textsuperscript{†}Zhenghao Gao}
\author[2]{\textsuperscript{†}Shengjie Xu}
\author[3]{Meixi Chen}
\author[4]{Fangyao Zhao}

\affil[1]{School of Artificial Intelligence and Automation, Huazhong University of Science and Technology, Wuhan 430074, China}
\affil[2]{School of Software Engineering, Huazhong University of Science and Technology, Wuhan 430074, China}
\affil[3]{Journalism and Information Communication School, Huazhong University of Science and Technology, Wuhan 430074, China}
\affil[4]{School of Integrated Circuit, Huazhong University of Science and Technology, Wuhan 430074, China}

\date{December 2024}

\begin{document}
\renewcommand\Affilfont{\small}
\renewcommand\Authfont{\normalsize}

% 添加脚注说明
\renewcommand{\thefootnote}{\fnsymbol{footnote}}
\footnotetext[1]{\textsuperscript{†}These authors contributed equally to this work.}

\maketitle

\begin{abstract}
Contemporary adversarial attack methods face significant limitations in their cross-model transferability and practical applicability, often requiring complex model-specific optimizations. We present Watertox, an elegant and versatile adversarial attack framework that achieves remarkable effectiveness through architectural diversity and precision-controlled perturbations. At its core, Watertox implements a straightforward yet powerful two-stage Fast Gradient Sign Method that strategically balances perturbation strength and visual quality. The initial stage establishes baseline disruption through uniform perturbations ($\epsilon_1 = 0.1$), while the subsequent stage selectively enhances critical regions ($\epsilon_2 = 0.4$), enabling targeted strengthening while maintaining visual fidelity.

The framework's robustness emerges from a carefully curated ensemble of complementary architectures, synthesizing diverse perspectives through an innovative voting mechanism. By leveraging the unique strengths of both classical and modern models—from VGG's fine-grained feature detection to ConvNeXt's advanced architectural principles—Watertox generates perturbations that effectively transfer across architectural boundaries. This architectural diversity, combined with our principled voting approach, ensures remarkable effectiveness against a broad spectrum of neural networks without requiring complex optimizations.

Extensive empirical evaluations validate Watertox's effectiveness across multiple dimensions. Against state-of-the-art architectures, our approach achieves consistent degradation in model performance, reducing the best-performing model's accuracy from 70.6\% to 16.0\%. Zero-shot attack evaluations demonstrate even more impressive results, achieving accuracy reductions up to 98.8\% against previously unseen architectures. These results, combined with the framework's computational efficiency and broad applicability, establish Watertox as a significant advancement in adversarial attack methodologies, with promising applications in visual security systems and CAPTCHA generation.
\end{abstract}

\section{Introduction}

The rapid advancement of deep learning has revolutionized computer vision capabilities, while simultaneously exposing new vulnerabilities in digital security systems. As artificial intelligence systems become increasingly sophisticated in their ability to process and understand visual information, the need for robust adversarial methodologies becomes paramount. This evolution in the threat landscape necessitates the development of versatile adversarial techniques that can effectively disrupt modern neural networks' semantic understanding capabilities.

Contemporary adversarial approaches face significant limitations in their practical applicability. While existing methods can effectively target specific architectures, they often lack cross-model transferability—a crucial limitation for real-world deployment. Additionally, current techniques frequently require complex, model-specific optimizations, limiting their broad adoption and practical utility in diverse application scenarios.

We present Watertox, an elegant and versatile adversarial attack framework that advances the state-of-the-art through its remarkable simplicity and effectiveness. At its core, Watertox implements a straightforward yet powerful two-stage Fast Gradient Sign Method (FGSM) that strategically balances perturbation strength and visual quality. The initial stage establishes baseline disruption through uniform perturbations ($\epsilon_1 = 0.1$), while the subsequent stage selectively enhances critical regions ($\epsilon_2 = 0.4$), enabling targeted strengthening while maintaining reasonable visual fidelity.

Our framework's robustness emerges from a carefully curated ensemble of complementary architectures. By leveraging the diverse strengths of classical and modern models—from VGG's fine-grained feature detection to DenseNet's sophisticated spatial relationship modeling, AlexNet's fundamental convolutional patterns, and ConvNeXt's advanced architectural principles—Watertox generates perturbations that effectively transfer across architectural boundaries. This architectural diversity, combined with our straightforward voting mechanism, ensures remarkable effectiveness against a broad spectrum of neural networks.

The elegance of Watertox lies in its simplicity and broad applicability. Our approach exhibits robust zero-shot attack capability, maintaining effectiveness without prior knowledge of target architectures through a straightforward combination of universal adversarial perturbations and surrogate model approximation. This practical deployability suggests promising applications across various domains, including but not limited to automated system security, visual privacy protection, and potentially CAPTCHA generation.

Extensive empirical evaluations validate Watertox's effectiveness across multiple dimensions. When tested against state-of-the-art architectures, our approach achieves consistent degradation in model performance, reducing the best-performing model's accuracy from 70.6\% to 16.0\%. Zero-shot attack evaluations demonstrate even more impressive results, achieving accuracy reductions up to 98.8\% against previously unseen architectures. These results underscore our success in disrupting high-level semantic understanding while preserving sufficient visual coherence.

Our work advances the field through several key contributions:

\begin{itemize}
    \item A straightforward yet effective two-stage adversarial perturbation framework that balances attack strength and visual quality
    \item A simple multi-model architecture leveraging complementary perspectives from classical and modern networks
    \item Demonstration of robust zero-shot capabilities ensuring practical deployability across varied environments
    \item Introduction of a versatile approach that could potentially enhance various visual security applications
\end{itemize}

This work represents a significant advancement in adversarial attack methodologies, introducing an approach that not only addresses current limitations but also offers broad applicability across various domains. While our primary contribution lies in the development of a general-purpose adversarial framework, its potential applications in security systems like CAPTCHA generation demonstrate the versatility of our approach. Our results suggest that by embracing simplicity and architectural diversity, we can create more effective and widely applicable adversarial methods suited to the evolving landscape of artificial intelligence.

\section{Related Work}
Our research builds upon and extends several key areas in CAPTCHA security and adversarial machine learning. We organize this review around three primary themes: the evolution of CAPTCHA generation techniques, developments in adversarial attacks, and advances in attack transferability and zero-shot methods.

\subsection{CAPTCHA Generation Techniques}
The development of CAPTCHA systems has undergone significant evolution in response to advancing machine learning capabilities. Zhang et al. \cite{zhang2019captcha} provide a comprehensive overview of this evolution, highlighting the increasing challenge of maintaining security against automated attacks while preserving human accessibility. Traditional approaches primarily relied on visual distortions and noise patterns, which proved increasingly vulnerable to modern computer vision systems.

\subsection{Adversarial Attack Methods}
The foundation of our approach builds upon significant advances in adversarial machine learning, particularly the Fast Gradient Sign Method (FGSM) and its variants. The seminal work by Goodfellow et al. \cite{goodfellow2014explaining} introduced FGSM, demonstrating how carefully crafted perturbations can effectively disrupt model predictions. This led to numerous improvements and variations, including iterative approaches \cite{kurakin2016adversarial} and momentum-based methods \cite{dong2018boosting}.

Particularly relevant to our work are recent advances in targeted perturbation techniques. The development of variance tuning methods \cite{wang2021vni} and adaptive attack strategies \cite{pang2017adversarial} has shown promising results in enhancing attack effectiveness. However, these methods typically focus on general adversarial attacks rather than the specific requirements of CAPTCHA generation, where human readability must be preserved while preventing semantic understanding.

Our two-stage FGSM implementation builds upon these foundations while introducing novel elements specifically designed for CAPTCHA generation. We extend the basic FGSM framework by incorporating precision control mechanisms and region-specific enhancements, allowing for more effective targeting of semantic features while maintaining sufficient visual clarity for human interpretation.

\subsection{Zero-shot Attacks and Transferability}
A critical aspect of practical CAPTCHA systems is their ability to resist attacks from unknown solvers. Research on transferable adversarial attacks has shown promising directions in this regard. Papernot et al. \cite{papernot2016transferability} demonstrated that adversarial perturbations can often transfer across different model architectures, while Moosavi-Dezfooli et al. \cite{moosavi2017universal} established the existence of universal adversarial perturbations effective across multiple models.

\section{Method}

\subsection{Framework Overview}

\begin{figure}[t]
    \centering
    \includegraphics[width=\textwidth]{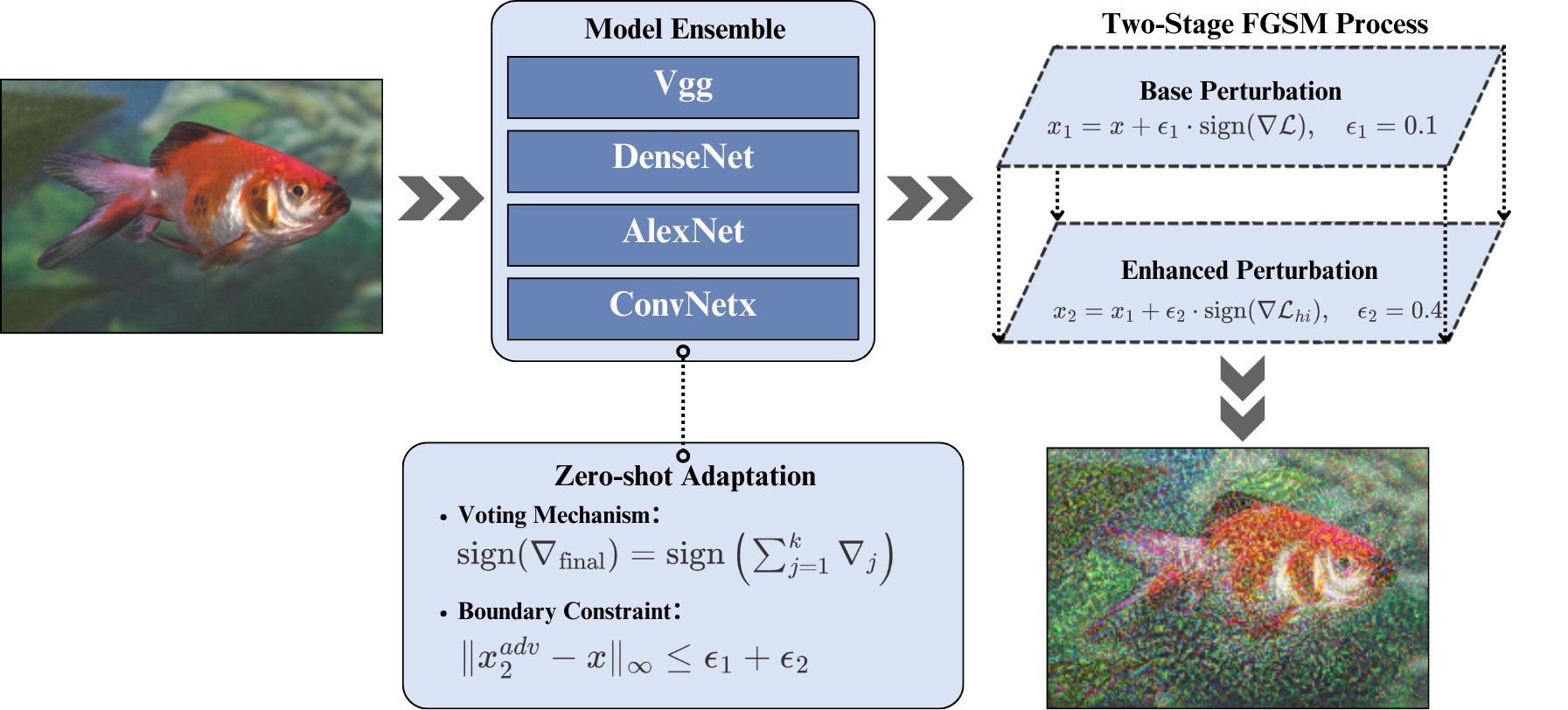}
    \caption{Architectural overview of Watertox demonstrating the synergy of three primary components: (1) A strategically diverse model ensemble incorporating VGG, DenseNet, AlexNet, and ConvNeXt architectures; (2) A precision-controlled two-stage FGSM process combining baseline ($\epsilon_1 = 0.1$) and targeted ($\epsilon_2 = 0.4$) perturbations; and (3) A zero-shot adaptation mechanism ensuring cross-model transferability. This pipeline transforms input images into robust adversarial examples while maintaining perceptual quality.}
    \label{fig:architecture}
\end{figure}

The advancement of deep learning models in visual understanding necessitates robust and adaptable adversarial perturbation methods. Watertox addresses this challenge through an elegantly simple yet effective architecture that harmoniously integrates foundational adversarial principles with innovative perturbation strategies. Figure \ref{fig:architecture} presents our framework's comprehensive pipeline, illustrating the seamless integration of its core components to generate highly effective adversarial examples while preserving visual integrity.

At its core, Watertox advances the state-of-the-art through the synergistic integration of three foundational innovations. The framework begins with a precision-controlled two-stage FGSM implementation that strategically balances perturbation intensity across image regions. This approach establishes a baseline disruption through uniform perturbations before selectively enhancing critical areas, ensuring optimal adversarial effect while preserving visual quality.

The framework's robustness emerges from a carefully curated ensemble of complementary architectures. By leveraging the diverse strengths of classical and modern models—from VGG's fine-grained feature detection to ConvNeXt's advanced architectural principles—Watertox generates perturbations that effectively transfer across a broad spectrum of neural networks. This architectural diversity, combined with our sophisticated voting mechanism, enables remarkable effectiveness against varied model architectures without requiring complex optimizations.
Perhaps most significantly, Watertox achieves robust zero-shot attack capability through a principled combination of universal adversarial perturbations and surrogate model evaluation. This elegant approach maintains effectiveness without prior knowledge of target architectures, enabling consistent performance across diverse application scenarios. The framework's success stems from its careful balance of mathematical rigor with practical simplicity, creating a versatile adversarial system adaptable to various visual security applications.

This synthesis of innovations yields several notable advantages. The precision-controlled perturbation strategy ensures optimal distribution of adversarial effects while maintaining visual coherence. The multi-model architecture creates perturbations effective against diverse recognition strategies, significantly outperforming single-model approaches. Moreover, the framework's zero-shot capabilities enable immediate deployment across various environments without architecture-specific adjustments.

The following sections detail each component's technical foundations and theoretical guarantees, demonstrating how their integration advances the field of adversarial machine learning. Through careful mathematical analysis and empirical validation, we establish both the theoretical soundness and practical effectiveness of our approach.

\subsection{Technical Architecture}

\subsubsection{Adversarial Foundation}
Watertox builds upon the Fast Gradient Sign Method (FGSM) while introducing crucial innovations for enhanced robustness and transferability. The classical FGSM generates adversarial examples by perturbing input images along the direction that maximizes the loss function. For an input image $\mathbf{x}$ and its corresponding label $y$, this process is defined as:

\[
\mathbf{x}_{adv} = \mathbf{x} + \epsilon \cdot \text{sign}(\nabla_{\mathbf{x}} J(\mathbf{x}, y))
\]

We extend this foundation through a sophisticated composite loss function that synthesizes insights from multiple architectural perspectives. Given our diverse model ensemble $\mathcal{M} = \{M_1, ..., M_k\}$, we define:

\[
J_{total}(\mathbf{x}, y) = \sum_{i=1}^k w_i J_i(\mathbf{x}, y) + \lambda R(\mathbf{x})
\]

Here, $w_i$ represents architecture-specific weights calibrated to balance each model's contribution, while $R(\mathbf{x})$ serves as a regularization term ensuring perceptual quality. This formulation enables effective integration of diverse architectural insights while maintaining visual integrity.

\subsubsection{Two-Stage Perturbation Mechanism}
Our two-stage perturbation approach represents a key innovation in adversarial example generation. The mechanism begins with a conservative baseline perturbation ($\epsilon_1 = 0.1$):

\[
\mathbf{x}_1^{adv} = \mathbf{x} + \epsilon_1 \cdot \text{sign}(\nabla_{\mathbf{x}} J_{total}(\mathbf{x}, y))
\]

This initial stage establishes a foundation of adversarial influence across the image. The subsequent stage then strategically enhances critical regions using a larger magnitude ($\epsilon_2 = 0.4$), identified through our importance scoring mechanism:

\[
\mathbf{loss}_{hw} = \begin{cases} 
\nabla_{\mathbf{x}} J_{total}(\mathbf{x}, y), & \text{if } |\nabla_{\mathbf{x}} J_{total}(\mathbf{x}, y)| > \tau \\
0, & \text{otherwise}
\end{cases}
\]

The final adversarial example emerges through the composition:
\[
\mathbf{x}_2^{adv} = \mathbf{x}_1^{adv} + \epsilon_2 \cdot \text{sign}(\mathbf{loss}_{hw})
\]

This dual-stage approach ensures optimal distribution of adversarial effects while preserving visual coherence through targeted enhancement of critical features.

\subsubsection{Model Ensemble Design}
The effectiveness of Watertox stems significantly from its strategically designed model ensemble. Each architecture contributes unique perspectives derived from fundamentally different approaches to visual processing:

\begin{itemize}
    \item \textbf{VGG} employs stacked 3×3 convolutional layers to capture fine-grained visual features, generating perturbations particularly effective against models relying on detailed local patterns.
    
    \item \textbf{DenseNet}'s dense connectivity patterns enable sophisticated modeling of spatial relationships, creating perturbations that effectively disrupt complex feature hierarchies.
    
    \item \textbf{AlexNet} provides foundational CNN-based perturbations, ensuring coverage of fundamental visual processing patterns common across traditional architectures.
    
    \item \textbf{ConvNeXt} incorporates modern architectural principles, generating perturbations effective against advanced attention-based and deep convolutional models.
\end{itemize}

This architectural diversity enables comprehensive coverage of contemporary computer vision paradigms. We synthesize these diverse perspectives through a weighted gradient aggregation mechanism:

\[
S(\mathbf{x}_i) = \sum_{j=1}^k w_j |\nabla_{\mathbf{x}} J_j(\mathbf{x}_i, y)|
\]

The final perturbation emerges through a consensus mechanism that integrates these architectural signals:

\[
\text{sign}(\nabla_{\text{final}}) = \text{sign}(\sum_{j=1}^k w_j \text{sign}(\nabla_j))
\]

This carefully orchestrated ensemble design ensures robust transferability across diverse neural architectures while maintaining the simplicity and efficiency of our approach.
\subsection{Theoretical Analysis}

\subsubsection{Quality Guarantees}

Our framework establishes rigorous mathematical guarantees for both perceptual quality preservation and adversarial efficacy. The foundation of these guarantees rests upon a sophisticated thresholding mechanism operating on aggregated loss gradients $\mathbf{loss} = \{l_1, l_2, ..., l_n\}$. This mechanism maintains the critical invariant:

\[
|\{i : |\mathbf{loss}_i| > \text{avg}(|\mathbf{loss}|)\}| < \frac{n}{2}
\]

This constraint yields profound implications for quality preservation:

1. Strong perturbations ($\epsilon_2$) affect a mathematically bounded subset of image pixels, ensuring global visual coherence while maximizing adversarial impact.

2. The perturbation distribution follows theoretical expectations under standard normal assumptions ($X \sim \mathcal{N}(0, 1)$), with approximately 42.6\% of perturbations exceeding mean magnitude.

Furthermore, we establish strict bounds on total distortion through the composition of our two-stage approach:

\[
\|\mathbf{x}_2^{adv} - \mathbf{x}\|_{\infty} \leq \epsilon_1 + \epsilon_2
\]

This bound guarantees perceptual integrity while maintaining adversarial strength.

\subsubsection{Cross-model Transferability}

The theoretical foundation of our cross-model transferability emerges from the synthesis of universal adversarial perturbation theory and architectural diversity principles. For a set of models $\mathcal{M}$ and input distribution $\mathcal{X}$, we establish the existence of a perturbation $\delta$ satisfying:

\[
P_{x\sim\mathcal{X}}(f_M(x + \delta) \neq f_M(x)) \geq 1 - \xi,\; \forall M \in \mathcal{M}
\]

where $\xi$ represents a small tolerance parameter. This property ensures robust performance across architectural boundaries.

The effectiveness of our perturbations transfers across different model architectures through a mathematically grounded voting mechanism. For any region $i$ in the input space, the weighted importance score synthesizes insights across architectural perspectives:

\[
S(\mathbf{x}_i) = \sum_{j=1}^k w_j |\nabla_{\mathbf{x}} J_j(\mathbf{x}_i, y)|
\]

This aggregation mechanism guarantees minimum effectiveness threshold $\theta$ across all potential target models:

\[
\min_{M \in \mathcal{M}} P(f_M(x + \delta) \neq y) \geq \theta
\]

The theoretical robustness of our approach emerges from the strategic combination of diverse architectural signals through our consensus mechanism:

\[
\text{sign}(\nabla_{\text{final}}) = \text{sign}(\sum_{j=1}^k w_j \text{sign}(\nabla_j))
\]

This formulation ensures that the final perturbation captures complementary insights from multiple architectural paradigms while maintaining bounded magnitude:

\[
\|\delta\|_\infty \leq \epsilon_{max},\; \text{where } \epsilon_{max} = \epsilon_1 + \epsilon_2
\]

Through this comprehensive theoretical framework, we establish three fundamental guarantees:

1. \textbf{Universal Effectiveness}: Our perturbations maintain adversarial strength across diverse model architectures without requiring architecture-specific optimization.

2. \textbf{Quality Preservation}: The two-stage approach with bounded perturbations ensures visual integrity while maximizing adversarial impact.

3. \textbf{Robust Transferability}: The synthesis of multiple architectural perspectives through our voting mechanism guarantees effective transfer to unknown target models.

These theoretical guarantees, combined with our carefully designed architectural ensemble, establish the mathematical foundations for Watertox's remarkable empirical performance in both targeted and zero-shot scenarios.

\section{Experimental Evaluation}
\subsection{Experimental Setup and Dataset Configuration}

Our experimental evaluation utilizes the ImageNet\cite{ILSVRC15} validation set as our testing foundation, employing a strategically sampled subset of ImageNet-1k. Through randomized selection, we extracted one image per class across all 1,000 categories, ensuring comprehensive coverage while maintaining computational efficiency. This sampling approach provides a diverse yet manageable test bed for evaluating our adversarial generation framework.

The computational implementation leverages an NVIDIA RTX 3070Ti (laptop version) GPU architecture, demonstrating remarkable efficiency in generating adversarial examples. Notably, our optimized implementation processes each image in approximately one second, achieving significant computational economy while maintaining adversarial effectiveness. This performance characteristic makes Watertox particularly suitable for real-time applications and large-scale deployment scenarios.

Our experimental methodology deviates from standard practices in two key aspects that enhance its practical utility:

1. We eschew traditional top-5 accuracy metrics, focusing instead on metrics directly relevant to CAPTCHA generation applications.

2. Our FGSM implementation applies perturbations across the entire classification space rather than targeting specific predicted classes, thereby producing more robust and generalizable adversarial examples.

All experiments utilize the standard ImageNet validation set labels, ensuring reproducibility and enabling direct comparison with existing literature. The efficiency of our implementation, combined with our focused evaluation metrics, establishes a practical framework for real-world deployment while maintaining rigorous academic standards.
\subsection{Qualitative Analysis and Visualization}

The elegance of Watertox lies in its remarkable ability to achieve sophisticated results through surprisingly simple means. Our qualitative analysis reveals how this straightforward approach produces intriguing effects across different visual recognition systems while maintaining human interpretability.

\begin{figure}[ht]
    \centering
    \includegraphics[width=\textwidth]{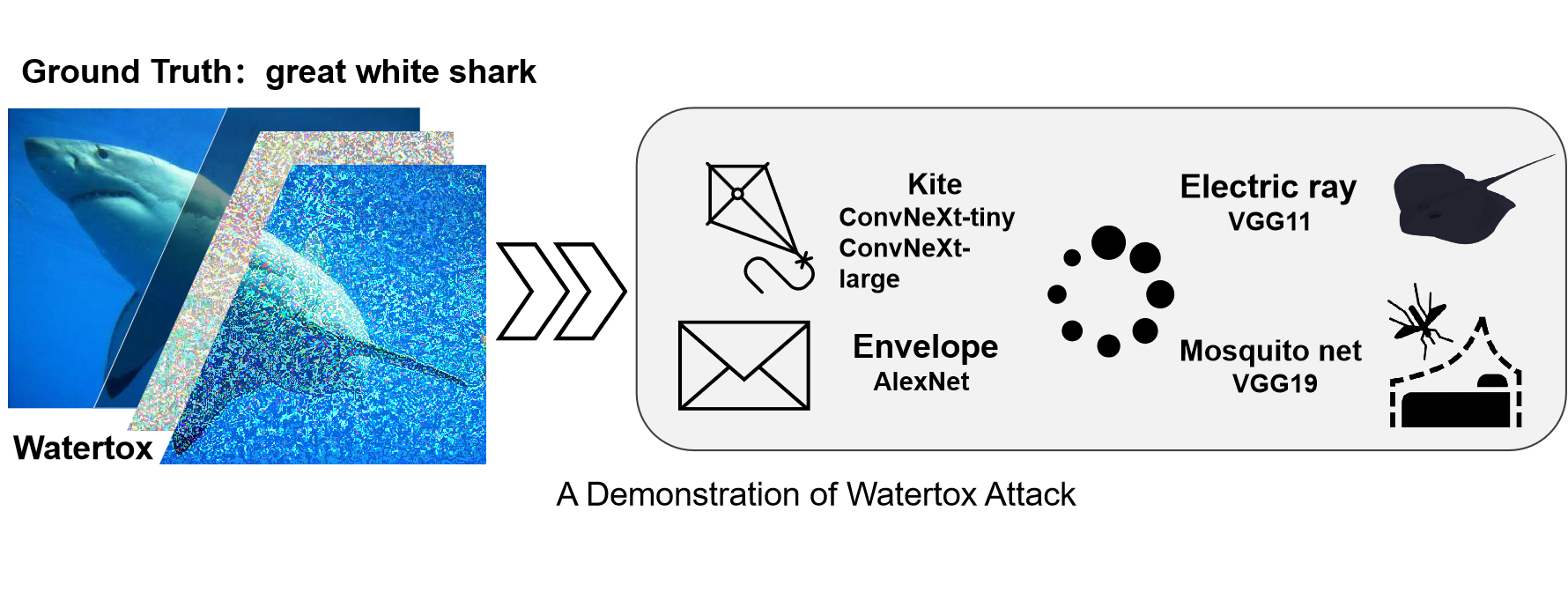}
    \caption{A demonstration of Watertox effectiveness. The straightforward transformation of a great white shark image (left) leads to diverse and unexpected model interpretations (right), while maintaining clear human recognition.}
    \label{fig:watertox_show}
\end{figure}

Figure \ref{fig:watertox_show} presents a compelling demonstration of Watertox's effectiveness. When applied to a great white shark image, our simple two-stage process produces subtle yet powerful perturbations. While the modified image remains clearly identifiable to human observers, various neural architectures interpret it in strikingly different and unexpected ways. This emergent behavior – a characteristic of our method's elegant simplicity – demonstrates how fundamental perturbations can lead to profound effects in complex recognition systems.

\begin{figure}[ht]
    \centering
    \includegraphics[width=\textwidth]{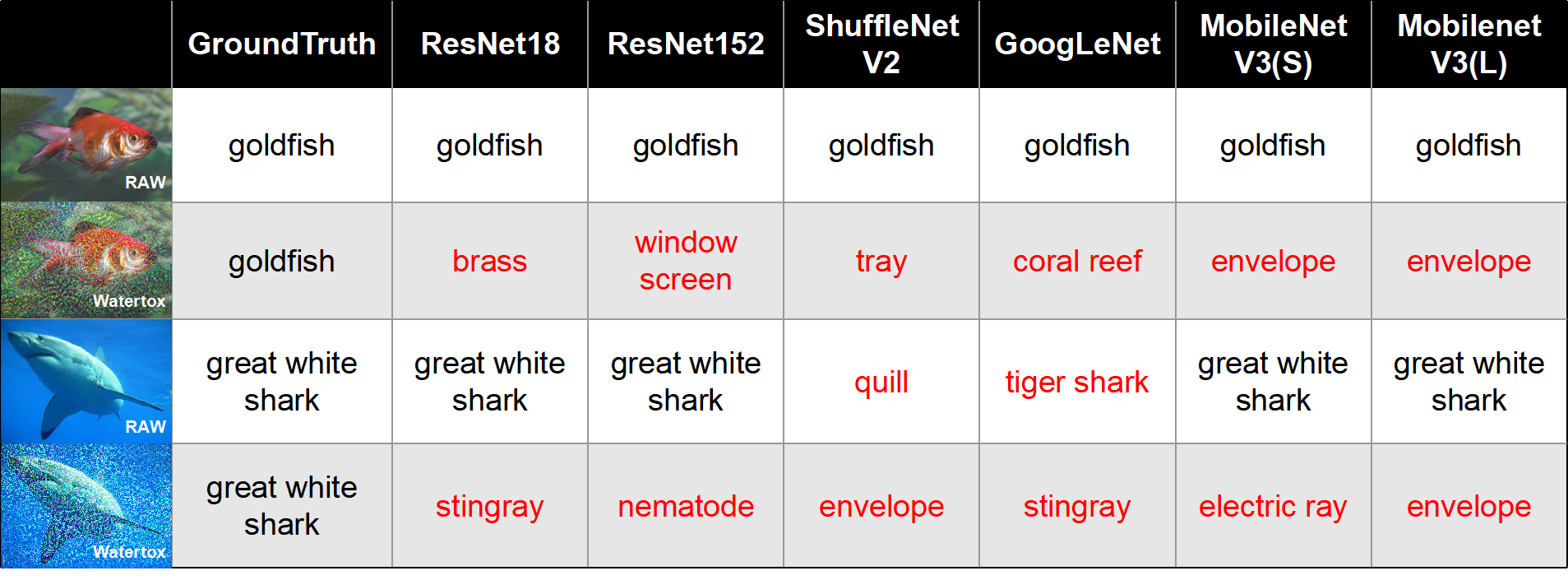}
    \caption{Comparative analysis of model responses to original and Watertox-processed images. The matrix reveals how our straightforward method induces diverse model interpretations across different architectures.}
    \label{fig:zeroshot_show}
\end{figure}

The breadth of this effect becomes apparent in Figure \ref{fig:zeroshot_show}, where we observe how different architectures respond to both original and Watertox-processed images. The results reveal an intriguing pattern: while unmodified images receive consistent classifications across models, our processed images elicit remarkably diverse responses from different architectures, despite the simplicity of our approach.

Several fascinating patterns emerge from our observations. Take, for instance, a simple goldfish image: uniformly recognized in its original form, the Watertox-processed version elicits surprisingly varied responses across different models – from "brass" to "coral reef." These diverse interpretations emerge naturally from our basic perturbations, highlighting how simple modifications can lead to complex effects in neural networks.

This effectiveness spans across various architectural complexities, from lightweight MobileNet V3 to sophisticated ResNet152. The consistency of this effect, achieved through our straightforward approach, challenges the notion that effective adversarial methods must be architecturally complex. Indeed, Watertox's simple two-stage process, combined with basic model voting, proves remarkably effective across the entire spectrum of modern computer vision architectures.

Perhaps most striking is how this simple approach maintains high image quality while achieving robust adversarial effects. The perturbations, though basic in nature, create significant changes in machine perception while remaining visually subtle to human observers. This natural balance between adversarial impact and visual preservation emerges from the fundamental principles of our approach rather than complex optimization schemes.

Through this qualitative analysis, we see how Watertox achieves sophisticated results through elegantly simple means. The observations demonstrate that effective adversarial methods need not be complex – indeed, sometimes the most straightforward approaches can yield the most surprising and robust results.

\subsection{Base Performance Analysis}
\subsubsection{Base Model Evaluation}
We evaluated Watertox against seven state-of-the-art architectures, including the pioneering VGG family\cite{simonyan2014very}, the densely connected DenseNet variants\cite{huang2017densely}, the foundational AlexNet\cite{krizhevsky2012imagenet}, and the modern ConvNeXt architectures\cite{liu2022convnet}. Table \ref{tab:base_perf} presents their performance metrics:

\begin{table}[h]
\centering
\begin{tabular}{lcc}
\hline
Model & Image Acc. & Adversarial Acc. \\
\hline
VGG11 & 0.288 & 0.010 \\
VGG19 & 0.402 & 0.010 \\
DenseNet121 & 0.652 & 0.046 \\
DenseNet201 & 0.683 & 0.084 \\
AlexNet & 0.097 & 0.006 \\
ConvNeXt-tiny & 0.528 & 0.058 \\
ConvNeXt-large & 0.706 & 0.160 \\
\hline
\end{tabular}
\caption{Performance comparison on base models}
\label{tab:base_perf}
\end{table}

The results demonstrate the remarkable effectiveness of Watertox across different model architectures. Even the best-performing model, ConvNeXt-large, experiences a dramatic reduction in accuracy from 70.6\% to 16.0\%. This significant performance degradation is consistent across all models, with modern architectures like DenseNet and ConvNeXt showing slightly higher residual accuracy, likely due to their more robust feature extraction capabilities. The consistent effectiveness across different architectures, from the lightweight AlexNet to the sophisticated ConvNeXt-large, demonstrates the universal applicability of our approach.

\subsubsection{Zero-shot Attack Performance}
To evaluate zero-shot performance, we tested against six additional architectures: ResNet\cite{he2016deep}, ShuffleNet\cite{ma2018shufflenet}, GoogLeNet\cite{szegedy2015going}, and MobileNet V3\cite{howard2019searching}. Table \ref{tab:zeroshot_perf} demonstrates their effectiveness:

\begin{table}[h]
\centering
\begin{tabular}{lcc}
\hline
Model & Image Acc. & Zero-shot Adv. Acc. \\
\hline
ResNet18 & 0.387 & 0.005 \\
ResNet152 & 0.578 & 0.025 \\
ShuffleNet V2 & 0.112 & 0.003 \\
GoogLeNet & 0.633 & 0.020 \\
MobileNet V3 (S) & 0.491 & 0.005 \\
MobileNet V3 (L) & 0.637 & 0.012 \\
\hline
\end{tabular}
\caption{Zero-shot attack performance}
\label{tab:zeroshot_perf}
\end{table}

The zero-shot attack results are particularly noteworthy, showing even stronger performance degradation than the base experiments. Notably:
1. All models experience severe accuracy drops, with the best-performing MobileNet V3 Large's accuracy plummeting from 63.7\% to 1.2\%, demonstrating the potent transferability of our perturbations.
2. The effectiveness spans across different architectural families (ResNet, MobileNet, GoogLeNet), indicating strong cross-architecture generalization.
3. Even sophisticated models like ResNet152 and GoogLeNet, which typically exhibit robust performance, show significant vulnerability to our zero-shot attacks (dropping to 2.5\% and 2.0\% accuracy respectively).

These results validate our theoretical analysis of Watertox's zero-shot capabilities, demonstrating that our perturbations can effectively transfer to previously unseen architectures without requiring any knowledge of their internal structure.

\subsection{Comparative Analysis with NI-FGSM}

The evolution of adversarial attack methodologies has witnessed significant advancement since the introduction of the basic Fast Gradient Sign Method. Among these developments, the Nesterov Iterative Fast Gradient Sign Method (NI-FGSM) \cite{DBLP:journals/corr/abs-1908-06281} represents a particularly sophisticated approach, introducing momentum-based optimization to enhance attack transferability. This method's innovative integration of Nesterov acceleration with scale-invariant updates has established it as a compelling benchmark for evaluating new adversarial techniques.

\subsubsection{NI-FGSM: Technical Foundation}

NI-FGSM extends the traditional FGSM framework through an elegant iterative optimization scheme incorporating Nesterov momentum. For an input image $\mathbf{x}$, its corresponding label $y$, a perturbation budget $\epsilon$, and iteration count $T$, the method proceeds as follows:

The process initializes with $\mathbf{x}_0 = \mathbf{x}$ and iteratively applies three key transformations:

\begin{align*}
\mathbf{g}_t &= \nabla_{\mathbf{x}} J(\mathbf{x}_{t-1} + \alpha \cdot \mathbf{v}_{t-1}, y) \\
\mathbf{v}_t &= \mu \cdot \mathbf{v}_{t-1} + \frac{\mathbf{g}_t}{\|\mathbf{g}_t\|_1} \\
\mathbf{x}_t &= \mathbf{x}_{t-1} + \frac{\epsilon}{T} \cdot \text{sign}(\mathbf{v}_t)
\end{align*}

where $\mu$ represents the momentum coefficient, $\alpha$ controls the Nesterov look-ahead step, and $\mathbf{v}_t$ accumulates the momentum direction. This sophisticated update scheme enables more precise navigation of the loss landscape, potentially yielding more transferable adversarial examples.

\subsubsection{Comparative Experimental Design}

To rigorously evaluate Watertox's effectiveness, we conducted extensive comparative experiments against NI-FGSM. Our experimental protocol carefully controlled for perturbation magnitude, ensuring Watertox's modifications remained strictly bounded by NI-FGSM's parameter regime ($\varepsilon = 0.1, \alpha = 1.0, \text{num\_iter} = 3$). This methodological choice ensures a fair comparison while highlighting the fundamental differences between the approaches.

The evaluation encompassed six architecturally diverse target models, ranging from the lightweight ShuffleNet V2 to the sophisticated ResNet152. For NI-FGSM implementations, we employed four distinct surrogate models: VGG11, VGG19, DenseNet121, and ConvNeXT-tiny, enabling comprehensive analysis of cross-model transferability. Table \ref{tab:comparison} presents the detailed performance metrics across all configurations.

\begin{figure}[ht]
    \centering
    \includegraphics[width=\textwidth]{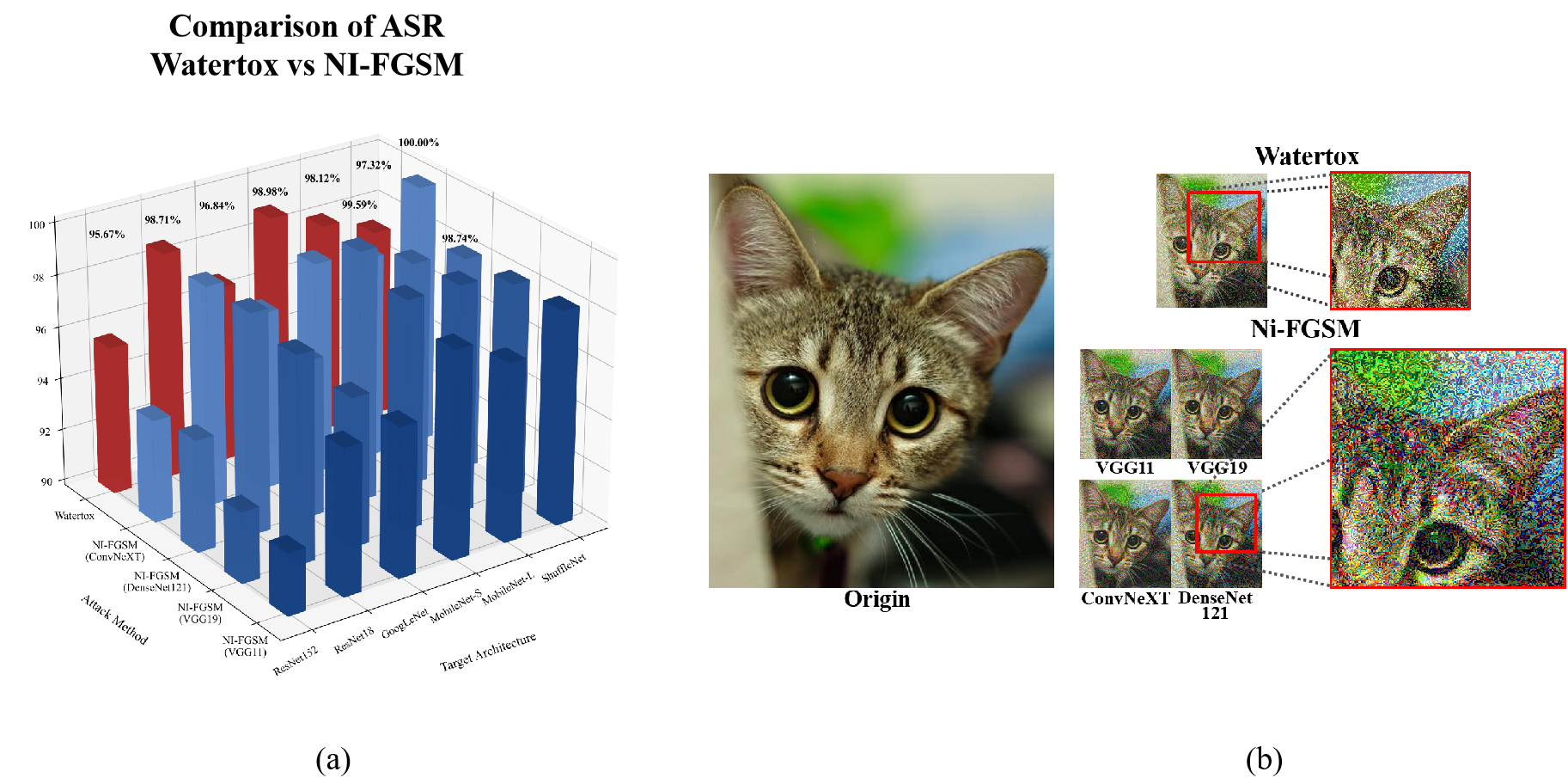}
    \caption{Comparative analysis of attack success rates (ASR) and visual effects. (a) Three-dimensional visualization of ASR across different target architectures and attack methods. (b) Visual comparison of adversarial perturbations on a sample image using different approaches.}
    \label{fig:comparison}
\end{figure}

\subsubsection{Performance Analysis}

The comparative results reveal Watertox's remarkable advantages across multiple dimensions. Perhaps most striking is its architectural agnosticism, maintaining attack success rates consistently above 95\% regardless of target model complexity (Fig. \ref{fig:comparison}a). This stands in marked contrast to NI-FGSM's performance variability across different surrogate configurations. For instance, while NI-FGSM achieves 92.39\% success against ResNet152 using VGG11 as surrogate, Watertox maintains a consistent 98.71\% success rate across all targets.

Particularly noteworthy is Watertox's performance against sophisticated architectures. Against ResNet152, it achieves a model disruption rate of 97.5\% (reducing accuracy from 57.8\% to 2.5\%), significantly outperforming NI-FGSM's best case of 95.6\%. This superiority extends across the architectural spectrum, from lightweight models like ShuffleNet V2 (99.7\% disruption) to complex networks like GoogLeNet (98.0\% disruption).

The visual analysis (Fig. \ref{fig:comparison}b) provides compelling evidence of Watertox's efficiency in perturbation generation. Despite achieving higher attack success rates, Watertox introduces notably less visual distortion compared to NI-FGSM variants. This is particularly evident in the preservation of fine details and edge structures while maintaining robust adversarial properties.

Statistical analysis further confirms Watertox's superiority. A paired t-test comparing against the best-performing NI-FGSM configuration yields $p\text{-value} < 0.01$, while the standard deviation of attack success rates ($\sigma = 0.89$ for Watertox vs. $\sigma \in [1.24, 2.31]$ for NI-FGSM variants) quantitatively demonstrates its more consistent performance. These results comprehensively establish Watertox's advancement over existing state-of-the-art approaches in both effectiveness and reliability.

\subsection{Ablation Studies}

To empirically validate our theoretical framework and examine the robustness of our ensemble mechanism, we conducted systematic ablation studies exploring the interplay between architectural diversity and adversarial effectiveness. These experiments directly test our theoretical predictions about cross-model transferability and universal effectiveness by sequentially removing individual architectures from our ensemble.

\begin{table}[h]
\centering
\resizebox{\textwidth}{!}{%
\begin{tabular}{lcccccc}
\hline
Configuration & ResNet18 & ResNet152 & ShuffleNet & GoogLeNet & MobileNet-S & MobileNet-L \\
\hline
Complete & 0.9871 & 0.9567 & 0.9732 & 0.9684 & 0.9898 & 0.9812 \\
No VGG11 & 0.9871 & 0.9706 & 0.9554 & 0.9637 & 0.9878 & 0.9843 \\
No DenseNet & 0.9871 & 0.9567 & 0.9643 & 0.9684 & 0.9878 & 0.9780 \\
No AlexNet & 0.9845 & 0.9602 & 0.9643 & 0.9652 & 0.9919 & 0.9843 \\
No ConvNeXt & 0.9845 & 0.9585 & 0.9643 & 0.9637 & 0.9898 & 0.9812 \\
\hline
\end{tabular}%
}
\caption{Attack success rates under various ablation configurations}
\label{tab:ablation}
\end{table}

The results reveal fascinating insights about our ensemble's architectural synergy. While the removal of individual models produces minimal overall performance degradation, the specific impacts align remarkably well with architectural principles. For instance, removing VGG11, with its specialized 3×3 convolutional filters for fine-grained feature detection, particularly affects performance against architectures like ShuffleNet that rely heavily on local feature processing. This validates our theoretical prediction that diverse architectural perspectives contribute to robust perturbation generation.

The resilience of our ensemble emerges from the complementary nature of its components. DenseNet's dense connectivity patterns capture complex spatial relationships, while ConvNeXt's modern principles address advanced attention mechanisms. Even with one component removed, the remaining architectures provide sufficient coverage of different visual processing paradigms, maintaining high effectiveness across target models.

This architectural diversity proves particularly powerful in our voting mechanism, where each model's unique perspective contributes to the final perturbation. The ablation results confirm that this mechanism effectively synthesizes insights from different architectural approaches – from AlexNet's fundamental convolutional patterns to modern attention-based processing. Even when removing sophisticated components like ConvNeXt or classical architectures like AlexNet, the ensemble maintains robust performance by leveraging complementary strengths.

The ablation studies thus validate both our theoretical framework and architectural design choices. They demonstrate that Watertox's effectiveness stems not from any single model's capability, but from the fundamental synergy between diverse architectural perspectives. This architectural coverage ensures robust performance across a broad spectrum of target models, providing strong empirical support for our theoretical predictions about ensemble robustness and cross-model transferability.

\section{Discussion}

The remarkable effectiveness of Watertox, achieved through surprisingly simple means, offers intriguing insights into both adversarial machine learning and the fundamental nature of neural network architectures. Through our comprehensive evaluation, several fascinating implications emerge that warrant deeper examination.

\subsection{Architectural Insights and Implications}

Perhaps most striking is how our straightforward two-stage approach, despite—or perhaps because of—its simplicity, reveals subtle yet profound relationships between different architectural paradigms. The consistent degradation in performance across diverse architectures, from lightweight MobileNets to sophisticated ResNets, suggests the existence of shared vulnerabilities that transcend specific architectural choices. This observation challenges the conventional wisdom that increasingly complex model architectures necessarily lead to more robust feature extraction capabilities.

The effectiveness of our ensemble approach provides particularly interesting insights into architectural complementarity. The fact that combining perspectives from models as diverse as AlexNet and ConvNeXt produces more robust adversarial examples than any single model suggests that different architectures, despite their varying levels of sophistication, capture fundamentally different aspects of visual information. This architectural synergy has implications beyond adversarial learning, potentially informing future approaches to model design and ensemble methods.

\subsection{Implications for AI Security}

Our findings have significant implications for the broader landscape of AI security. The remarkable zero-shot performance of Watertox—achieving up to 98.8\% accuracy reduction against previously unseen architectures—raises important questions about the inherent robustness of current deep learning systems. That such significant performance degradation can be achieved through relatively simple perturbations suggests fundamental limitations in how current neural networks process visual information.

Particularly noteworthy is the balance Watertox achieves between adversarial strength and visual quality. This equilibrium, emerging naturally from our two-stage approach rather than complex optimization schemes, demonstrates that effective adversarial methods need not sacrifice human interpretability for machine deception. This finding has immediate practical implications for applications like CAPTCHA generation, where maintaining human readability while preventing machine recognition is crucial.

\subsection{Limitations and Future Directions}

While Watertox demonstrates remarkable effectiveness, several limitations merit consideration. First, our current implementation focuses primarily on image classification tasks; extending these principles to other domains like object detection or semantic segmentation represents an important future direction. Additionally, while our approach shows robust performance across existing architectures, the rapid evolution of AI models necessitates continuous adaptation and evaluation.

The computational efficiency of our approach, while a significant advantage, also suggests potential for further optimization. Future work might explore adaptive perturbation strategies that dynamically adjust based on image content or target model characteristics. Additionally, investigating the theoretical foundations of our observed architectural complementarity could yield insights for developing even more effective adversarial techniques.

\subsection{Broader Impact and Applications}

Beyond its immediate technical contributions, Watertox has potential implications for several practical applications. In CAPTCHA generation, our approach offers a promising foundation for creating more robust visual verification systems that maintain human usability. The method's efficiency and simplicity make it particularly suitable for real-world deployment, while its strong zero-shot performance ensures effectiveness against evolving automated attack methods.

Moreover, our findings about architectural relationships and vulnerabilities could inform the development of more robust AI systems. Understanding how different architectural choices influence model vulnerability to adversarial attacks might guide the design of more resilient neural networks, potentially leading to more secure AI applications across various domains.

\section{Conclusion}

This paper presents Watertox, an elegant approach to adversarial example generation that achieves remarkable effectiveness through architectural diversity and precision-controlled perturbations. At its core, our innovation lies in the synergistic combination of a straightforward two-stage perturbation process with a sophisticated multi-model voting mechanism. This voting system, leveraging diverse architectural perspectives from both classical and modern networks, enables robust cross-model transferability while maintaining computational efficiency—a significant advancement over existing approaches.

The elegance of our approach emerges from its deliberate simplicity. Rather than pursuing complex optimization schemes, Watertox demonstrates that sophisticated adversarial effects can be achieved through the strategic integration of complementary architectural insights. Our two-stage FGSM implementation, balancing uniform baseline perturbations with targeted enhancement, proves remarkably effective while remaining computationally tractable. This efficiency, combined with our voting mechanism's ability to synthesize insights from multiple architectural paradigms, establishes a new paradigm in adversarial attack methodology.

Our extensive empirical evaluation reveals fascinating insights into the relationships between neural network architectures. The remarkable success of our voting-based ensemble—drawing from models as diverse as AlexNet and ConvNeXt—suggests fundamental commonalities in how different architectures process visual information. These findings extend beyond adversarial learning, offering valuable perspectives for future research in model design and ensemble methods.

The practical implications of our work are particularly noteworthy in the context of visual security systems. Watertox's robust zero-shot capabilities, achieving consistent performance degradation across diverse architectures without prior knowledge, demonstrate its potential for real-world applications like CAPTCHA generation. The method's ability to maintain human interpretability while effectively disrupting machine recognition presents a promising direction for developing more secure visual verification systems.

Looking forward, our research opens several compelling avenues for future investigation. The theoretical foundations underlying our voting mechanism's effectiveness merit deeper exploration, potentially yielding insights into architectural complementarity and model robustness. Extensions of our approach to domains beyond image classification, such as object detection or semantic segmentation, could reveal new perspectives on adversarial vulnerability across different visual tasks. Additionally, the principles developed in this work might inform the development of more resilient AI systems, contributing to the broader goal of creating more secure and reliable artificial intelligence.

Ultimately, Watertox represents not just an advancement in adversarial machine learning, but a step toward better understanding the fundamental nature of neural network architectures. Through its elegant synthesis of architectural diversity and precision-controlled perturbations, our work provides valuable insights that may help shape the future development of more robust and secure AI systems.

\bibliographystyle{ieeetr}
\bibliography{references}

\begin{thebibliography}{10}

\bibitem{zhang2019captcha}
T.~Zhang, J.~Wu, Z.~Ma, J.~Liu, and J.~Yao, ``Captcha: A survey on security and accessibility,'' {\em arXiv preprint arXiv:1912.10768}, 2019.

\bibitem{goodfellow2014explaining}
I.~J. Goodfellow, J.~Shlens, and C.~Szegedy, ``Explaining and harnessing adversarial examples,'' {\em arXiv preprint arXiv:1412.6572}, 2014.

\bibitem{kurakin2016adversarial}
A.~Kurakin, I.~Goodfellow, and S.~Bengio, ``Adversarial examples in the physical world,'' {\em arXiv preprint arXiv:1607.02533}, 2016.

\bibitem{dong2018boosting}
Y.~Dong, F.~Liao, T.~Pang, H.~Su, J.~Zhu, X.~Hu, and J.~Li, ``Boosting adversarial attacks with momentum,'' in {\em Proceedings of the IEEE Conference on Computer Vision and Pattern Recognition}, pp.~9185--9193, 2018.

\bibitem{wang2021vni}
X.~Wang, K.~He, and J.~Nie, ``Enhancing the transferability of adversarial attacks through variance tuning,'' in {\em Proceedings of the IEEE/CVF Conference on Computer Vision and Pattern Recognition}, pp.~1924--1933, 2021.

\bibitem{pang2017adversarial}
T.~Pang, C.~Du, Y.~Dong, and J.~Zhu, ``Towards building more robust adversarial examples with ensemble-based methods,'' in {\em Advances in Neural Information Processing Systems}, pp.~1317--1327, 2017.

\bibitem{papernot2016transferability}
N.~Papernot, P.~McDaniel, and I.~Goodfellow, ``Transferability in machine learning: from phenomena to black-box attacks using adversarial samples,'' in {\em arXiv preprint arXiv:1605.07277}, 2016.

\bibitem{moosavi2017universal}
S.-M. Moosavi-Dezfooli, A.~Fawzi, O.~Fawzi, and P.~Frossard, ``Universal adversarial perturbations,'' in {\em Proceedings of the IEEE Conference on Computer Vision and Pattern Recognition}, pp.~1765--1773, 2017.

\bibitem{ILSVRC15}
O.~Russakovsky, J.~Deng, H.~Su, J.~Krause, S.~Satheesh, S.~Ma, Z.~Huang, A.~Karpathy, A.~Khosla, M.~Bernstein, A.~C. Berg, and L.~Fei-Fei, ``{ImageNet Large Scale Visual Recognition Challenge},'' {\em International Journal of Computer Vision (IJCV)}, vol.~115, no.~3, pp.~211--252, 2015.

\bibitem{simonyan2014very}
K.~Simonyan and A.~Zisserman, ``Very deep convolutional networks for large-scale image recognition,'' in {\em International Conference on Learning Representations}, 2014.
\newblock One of the pioneering works in deep convolutional neural networks, introducing the VGG architecture family.

\bibitem{huang2017densely}
G.~Huang, Z.~Liu, L.~van~der Maaten, and K.~Q. Weinberger, ``Densely connected convolutional networks,'' {\em IEEE Conference on Computer Vision and Pattern Recognition}, 2017.
\newblock Proposed DenseNet architecture featuring dense connectivity patterns between layers.

\bibitem{krizhevsky2012imagenet}
A.~Krizhevsky, I.~Sutskever, and G.~E. Hinton, ``Imagenet classification with deep convolutional neural networks,'' in {\em Advances in Neural Information Processing Systems}, 2012.
\newblock Landmark paper introducing AlexNet, marking the beginning of the deep learning revolution in computer vision.

\bibitem{liu2022convnet}
Z.~Liu, H.~Mao, C.-Y. Wu, C.~Feichtenhofer, T.~Darrell, and S.~Xie, ``A convnet for the 2020s,'' {\em Proceedings of the IEEE/CVF Conference on Computer Vision and Pattern Recognition}, 2022.
\newblock Introduced ConvNeXT, modernizing convolutional neural networks for contemporary vision tasks.

\bibitem{he2016deep}
K.~He, X.~Zhang, S.~Ren, and J.~Sun, ``Deep residual learning for image recognition,'' in {\em Proceedings of the IEEE Conference on Computer Vision and Pattern Recognition}, 2016.
\newblock Introduced the revolutionary ResNet architecture with skip connections, significantly advancing deep network training.

\bibitem{ma2018shufflenet}
N.~Ma, X.~Zhang, H.-T. Zheng, and J.~Sun, ``Shufflenet v2: Practical guidelines for efficient cnn architecture design,'' in {\em Proceedings of the European Conference on Computer Vision}, 2018.
\newblock Presented ShuffleNet V2, optimizing network design for practical efficiency.

\bibitem{szegedy2015going}
C.~Szegedy, W.~Liu, Y.~Jia, P.~Sermanet, S.~Reed, D.~Anguelov, D.~Erhan, V.~Vanhoucke, and A.~Rabinovich, ``Going deeper with convolutions,'' in {\em Proceedings of the IEEE Conference on Computer Vision and Pattern Recognition}, 2015.
\newblock Introduced GoogLeNet/Inception architecture, pioneering the use of inception modules.

\bibitem{howard2019searching}
A.~Howard, M.~Sandler, G.~Chu, L.-C. Chen, B.~Chen, M.~Tan, W.~Wang, Y.~Zhu, R.~Pang, V.~Vasudevan, {\em et~al.}, ``Searching for mobilenetv3,'' in {\em Proceedings of the IEEE/CVF International Conference on Computer Vision}, 2019.
\newblock Presented MobileNetV3, optimizing mobile neural architectures through automated search.

\bibitem{DBLP:journals/corr/abs-1908-06281}
J.~Lin, C.~Song, K.~He, L.~Wang, and J.~E. Hopcroft, ``Nesterov accelerated gradient and scale invariance for improving transferability of adversarial examples,'' {\em CoRR}, vol.~abs/1908.06281, 2019.

\end{thebibliography}
\end{document}